\documentclass[conference]{IEEEtran}
\IEEEoverridecommandlockouts

\usepackage{amsmath,graphicx, multirow, hyperref}

\usepackage{cite}
\usepackage{amsmath,amssymb,amsfonts}
\usepackage{algorithmic}
\usepackage{graphicx}
\usepackage{textcomp}
\usepackage{xcolor}
\def\BibTeX{{\rm B\kern-.05em{\sc i\kern-.025em b}\kern-.08em
    T\kern-.1667em\lower.7ex\hbox{E}\kern-.125emX}}

\makeatletter
\newcommand{\linebreakand}{%
  \end{@IEEEauthorhalign}
  \hfill\mbox{}\par
  \mbox{}\hfill\begin{@IEEEauthorhalign}
}
\makeatother

\begin{document}

\title{ASR Benchmarking: Need for a More Representative Conversational Dataset}

\author{
    \IEEEauthorblockN{
        Gaurav Maheshwari*\thanks{*These three authors contributed equally}\IEEEauthorrefmark{2}, Dmitry Ivanov*\IEEEauthorrefmark{2}, Théo Johannet*\IEEEauthorrefmark{2}, Kevin El Haddad\IEEEauthorrefmark{2}\IEEEauthorrefmark{3}
    }
    \IEEEauthorblockA{\IEEEauthorrefmark{2} Diabolocom}
    \IEEEauthorblockA{\IEEEauthorrefmark{3} ISIA Lab - University of Mons}
    \IEEEauthorblockA{first\_name.last\_name@diabolocom.com}
}

\maketitle

\begin{abstract}
Automatic Speech Recognition (ASR) systems have achieved remarkable performance on widely used benchmarks such as LibriSpeech and Fleurs. However, these benchmarks do not adequately reflect the complexities of real-world conversational environments, where speech is often unstructured and contains disfluencies such as pauses, interruptions, and diverse accents. In this study, we introduce a multilingual conversational dataset, derived from TalkBank, consisting of unstructured phone conversation between adults. Our results show a significant performance drop across various state-of-the-art ASR models when tested in conversational settings. Furthermore, we observe a correlation between Word Error Rate and the presence of speech disfluencies, highlighting the critical need for more realistic, conversational ASR benchmarks.
\end{abstract}


\section{Introduction}
\label{sec:intro}

Automatic Speech Recognition (ASR) systems have made significant advancements in recent years, with models like Whisper~\cite{radford2023robust}, wav2vec2~\cite{baevski2020wav2vec}, and Kaldi~\cite{povey2011kaldi, ravanelli2019pytorch}  pushing the boundaries of ASR accuracy. For instance, Canary~\cite{puvvada2024less}, which uses a FastConformer-based~\cite{rekesh2023fast} encoder-decoder architecture, reports an impressive average Word Error Rate (WER) of 0.065 on the Hugging Face Open ASR leaderboard~\cite{open-asr-leaderboard}. These breakthroughs stem not only from improvements in model architectures but also from the availability of large, diverse datasets like CommonVoice~\cite{ardila2019common}, and LibriSpeech~\cite{panayotov2015librispeech} which facilitate pre-training and provide standardized benchmarks.

However, most existing ASR benchmarks are derived in controlled, and non-conversational settings. For example, LibriSpeech is sourced from audiobook recordings, while Fleurs~\cite{conneau2023fleurs} is collected in highly controlled environments with native speakers. As a result, ASR models trained and benchmarked on these datasets may not perform well in real-world conversation oriented conditions, which often include speech disfluencies, diverse accents, and noisy backgrounds.

To address these limitations, we introduce a new benchmark based on the TalkBank~\cite{macwhinney2004talkbank} dataset—a multilingual, conversational corpus that captures real-world, unstructured speech. Our TalkBank subset consists of phone conversations between adults, with natural speech disfluencies like laughter, pauses, and interjections. We perform extensive preprocessing on this dataset, including channel separation, transcript alignment, and filtering out mismatched recordings, ensuring an accurate and reliable resource for conversational ASR benchmarking.

We evaluate multiple modern ASR systems on this challenging dataset and compare their performance with results on existing benchmarks. Our analysis reveals a substantial performance drop in conversational settings. For instance, while CANARY achieves a WER of 0.19 on LibriSpeech, its WER on TalkBank rises to 0.54. Other models also exhibit similar trends. Additionally, we analyze the impact of different speech disfluencies and find a correlation between WER and these disfluencies. Overall, our results indicate that ASR models still underperform in conversational environments, and current benchmarks do not adequately capture these challenges.

Our primary contributions are as follows:

\begin{itemize}
    \item We process TalkBank, a conversationally-oriented multilingual dataset consisting of unstructured conversations between adults, featuring real-world speech disfluencies.
    \item We benchmark various ASR models on this dataset, and find that conversational speech remains a significant challenge for existing ASR systems.
\end{itemize}

\section{Dataset}

In this section, we describe TalkBank, the set of preprocessing steps we applied, and some statistics of the final dataset.

\subsection{TalkBank}

TalkBank is a large publicly accessible database containing spoken language data that supports various areas of research, including speech-language pathology, language acquisition, and bilingualism~\cite{wong2023hierarchical}. The database is primarily composed of 14 different components, such as AphasiaBank~\cite{macwhinney2011aphasiabank}, which is intended for the study of language in aphasia, and PhonBank~\cite{rose2014phonbank}, which focuses on the analysis of children’s phonological development in 18 languages.

In this work, we focus on Conversation Banks (CABank)~\cite{macwhinney2010transcribing}, which was primarily developed for studying conversations between adults using the methods of conversation analysis. CABank consists of multiple corpora in different languages and settings, such as phone calls, meeting recordings, and lectures. Here, we focus on CallFriend and CallHome, both of which are recordings of phone conversations between two adults in various languages.

\subsection{Pre-processing}

Both CallFriend and CallHome consist of audio recordings and corresponding transcripts. However, upon manual inspection, we found that in several cases, the audio segments did not align with the given transcripts. Moreover, the dataset lacks information about speaker channels. Additionally, since the transcripts are in CHAT format\footnote{https://talkbank.org/manuals/CHAT.html}, they cannot be directly compared with the output of the ASR. Therefore, we applied a series of preprocessing steps to clean the dataset and align the format with the output of the ASR.

\begin{figure}
    \centering
    \includegraphics[width=0.8\linewidth]{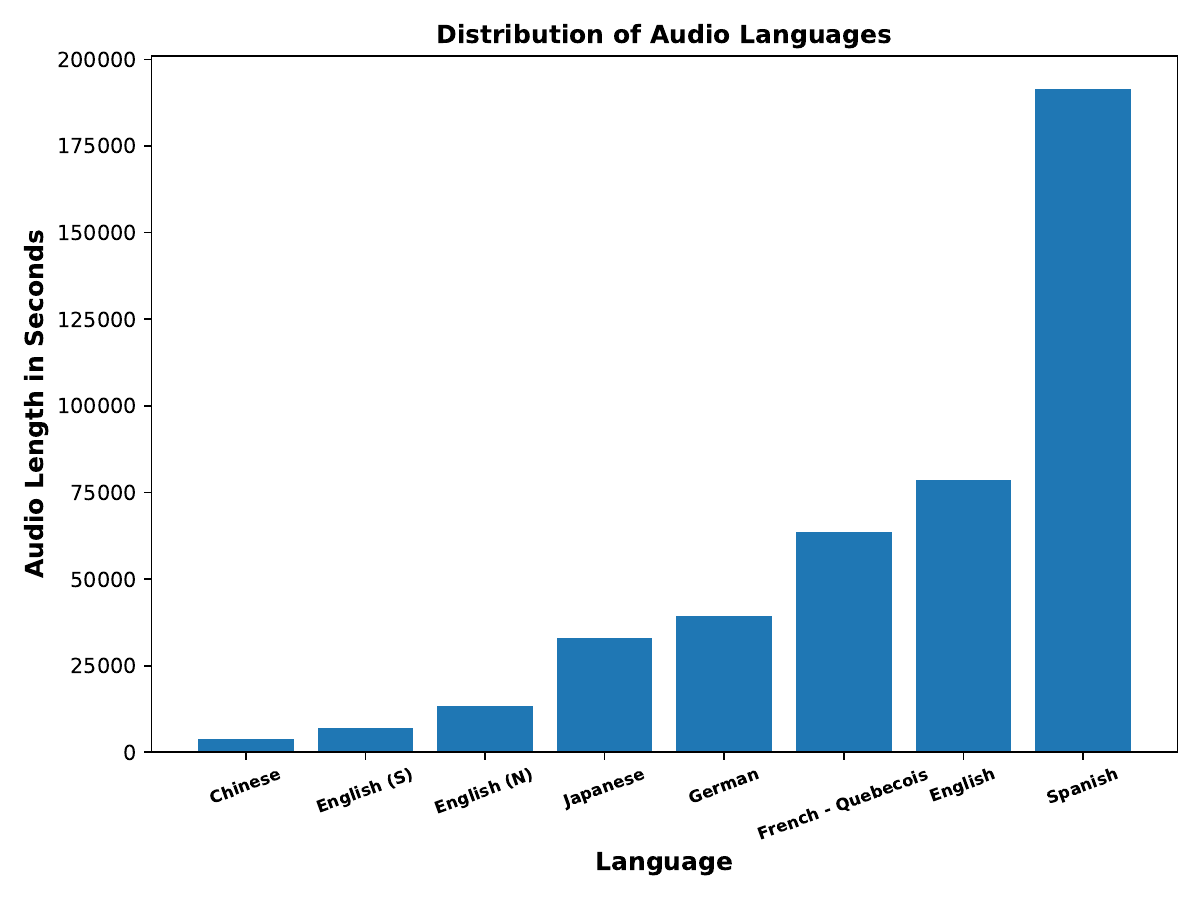}
    \caption{Distribution of audio length in seconds per language.}
    \label{fig:language-distribution}
\end{figure}
\paragraph{Manual Filtering} We began the preprocessing by manually reviewing the audio files and their associated transcripts. We removed all files where (i) the transcripts and audio were completely different, or (ii) there was only one speaker.

\paragraph{Aligning speaker with channel} The TalkBank corpus does not contain information about the channel associate with speakers. To address this, we employed a Voice Activity Detection (VAD) based method to accurately map speakers to their respective audio channels. We segmented the audio into intervals based on the transcript timestamps and analyzed each segment for speech activity across all channels. More specifically, for each segment, we calculated the percentage of speech detected by pyannote~\cite{Plaquet23, Bredin23} based VAD for each channel. We then averaged these speech presence percentages across all segments of the audio file to determine which channel contained a higher concentration of the speaker's voice, thereby identifying the primary channel for each speaker. We discarded all audio files where we could not accurately determine speaker channel based on speech concentration.


\paragraph{Removing annotation with no time stamp} Once the speaker channels were identified, we proceeded to filter out annotations that lacked timestamps. Additionally, we observed that the first annotation following a non-timestamped entry often included text that belonged to the previous non-timestamped annotation. To ensure the accuracy of our data, we also removed the annotation immediately following any non-timestamped entry.

\paragraph{Aligning time stamp with VAD} During manual examination, we found that some annotation timestamps were not properly aligned with the corresponding speech segments, often ending too soon resulting in cutting off speech prematurely. To address this, we utilized a VAD model to accurately detect speech segments. If an annotation's end timestamp fell within a detected speech segment, we extended the end timestamp to match the end of the speech as identified by the VAD model. However, to prevent overlap, we limited the new end timestamp to the start timestamp of the next annotation. This ensured accurate alignment between annotations and speech while avoiding conflicts between consecutive annotations.

\paragraph{Discarding Audio Segments} Despite discarding audio files that did not align with transcripts and applying extensive preprocessing, we still found audio segments where the transcripts were either prematurely cut off or entirely inaccurate. Given that manual inspection was impractical for over 100,000 audio segments, we developed an ASR models based automated filtering process. We discard files where two different ASR models produced outputs similar to each other but both exhibited a high WER. The rationale being that two distinct ASR models are unlikely to produce identical error patterns and is likely an issue at the transcript level rather than with the models.


Formally, consider audio segment $A$, with the corresponding transcript $t_{A}$. Additionally, consider two ASR models $ASR_1$ and $ASR_2$ with their corresponding out output as $t_{ASR_1}$ and $t_{ASR_2}$. We define rejection factor $k$ as:

\begin{equation*}
    \begin{split}
         k = abs(&\frac{wer(t_{ASR_1}, t_{ASR_2}) + wer(t_{ASR_2}, t_{ASR_1})}{2} - \\ &\frac{wer(t_{A}, t_{ASR_1}) + wer(t_{A}, t_{ASR_2})}{2})
    \end{split}
\end{equation*}

Here $wer(\cdot)$ refers to the Word Error Rate. We discarded an audio segment if the $k$ exceeded a certain threshold. A very low $k$ would overly favor ASR models, while a very high $k$ could result in retaining incorrect transcripts, thereby compromising data quality. To balance this, we manually reviewed discarded audio files across different $k$ values and determined that a $k$ of $0.8$ provided the optimal balance, effectively filtering out erroneous transcripts without favouring ASR models.

\paragraph{Final Transcripts} The transcripts from TalkBank are provided in the CHAT format, which includes various special symbols representing conversational events such as laughter, pauses, and non-verbal expressions. These symbols prevent direct comparison between ASR model outputs and the transcripts. To address this, we preprocess the transcripts by replacing these special symbols with empty strings. The replacements are available in the source code and are based on guidelines from the CHAT manual. This step ensures that the ASR model outputs can be compared with the transcripts.

\paragraph{Speaker Switch Version} As an alternative to segmenting based on the provided annotations, which we refer to as Talkbank Segments, we explored a different approach where we merged all consecutive segments attributed to the same speaker into a single, larger audio segment. This method resulted in fewer, but longer, audio segments. In our experiments, we compared the performance of this speaker-based segmentation (TalkBank Switch) with the original annotation-based segmentation (TalkBank Segments) to evaluate its impact on model performance.


After applying the aforementioned preprocessing steps, the final dataset consists of $151,705$ audio segments across eight languages. Figure~\ref{fig:language-distribution} presents the distribution of data by the total duration of audio segments for each language. The dataset, including test-train splits and preprocessing details, is available in the source code repository\footnote{\url{https://github.com/Diabolocom-Research/ConversationalDataset}}. For the test-train split, we maintain the same proportion per language as used in the CommonVoice dataset.

\section{Experiments}

Recall that the primary objective of this study is to understand the performance of various ASR models in conversational settings. To that end, we evaluate the performance of several ASR models on the TalkBank dataset and compare their results against other standard reference datasets. We then analyze the impact of conversation-specific aspects, such as laughter and interruptions, on the WER of these ASR models. We begin by describing various models and reference datasets.

\begin{figure*}
    \centering
    \includegraphics[width=1\linewidth]{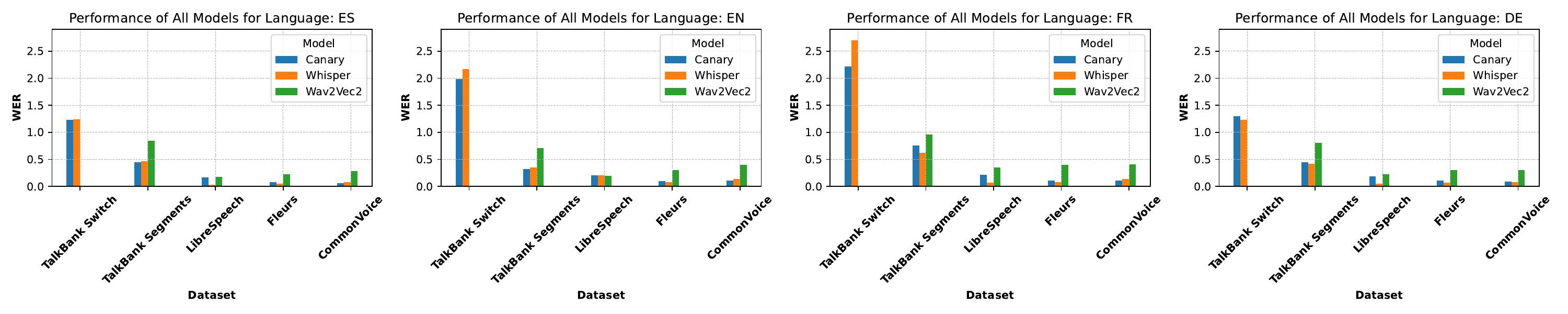}
    \caption{WER of various ASR system with respect to Spanish (ES), English (EN), French (FR), and German (DE).}
    \label{fig:wer-language}
\end{figure*}

\textbf{Models} In this study, we benchmark three families of open-source ASR models, namely: (i) \textbf{Whisper}, which consists of a transformer-based encoder-decoder architecture trained on a large multilingual dataset with 30-second audio segments. We use the Whisper Large V3 multilingual model with $1,550$M parameters, (ii) \textbf{wav2vec2}, which employs a self-supervised training regime that aims to predict the correct latent representation among several candidates, followed by a fine-tuning step for ASR. For our experiments, we use the wav2vec2-XLSR-Multilingual-56 model, and (iii) \textbf{Canary}, which has a similar architecture to Whisper but uses FastConformer instead of a transformer and employs dynamic batching instead of $30$-second audio segments. In our experiments, we use the 1B parameter multilingual model.

\begin{table}[]
\centering
\begin{tabular}{llllll}\hline
\multicolumn{1}{c}{\multirow{2}{*}{Method}} & \multicolumn{5}{c}{Dataset}        \\ \cline{2-6} 
\multicolumn{1}{c}{}                        & LS  & FL  & CV  & Segment & Switch \\ \hline
wav2vec2                                   & 0.22 & 0.42 & 0.39 & 0.86     & N/A    \\
Whisper                                     & 0.11 & 0.23 & 0.17 & 0.54     & 1.73    \\
Canary                                      & 0.19 & 0.25 & 0.16 & 0.54     & 1.37   \\
\hline\\
\end{tabular}
\caption{WER (lower is better) of ASR systems across multilingual LibriSpeech (LS), Fleurs (FL), CommonVoice (CV), TalkBank Segments, and TalkBank Switch.}
\label{tab:wer-overall}
\end{table}

\begin{table}[]
\centering
\begin{tabular}{lllll}\hline
\multicolumn{1}{c}{\multirow{2}{*}{Method}} & \multicolumn{4}{c}{Time}                                                                              \\ \cline{2-5} 
\multicolumn{1}{c}{}                        & T \textless{} 5 & 5 \textless T \textless{} 10 & 10 \textless T \textless{} 20 & T \textgreater 20 \\ \hline
wav2vec2                                   & 0.87              & 0.79                           & 0.81                            & 0.77               \\
Whisper                                     & 0.56              & 0.45                           & 0.50                            & 0.58               \\
Canary                                      & 0.55              & 0.39                           & 0.43                            & 0.54     \\ \hline \\        
\end{tabular}
\caption{WER (lower is better) of ASR systems on TalkBank Segments based on different time duration: Audio segments shorter than 5 seconds (T\textless{}5), between 5 and 10 seconds (5 \textless T \textless{} 10), between 10 and 20 seconds (10 \textless T \textless{} 20), and longer than 20 seconds (T \textgreater 20).}
\label{tab:wer-time}
\end{table}

\textbf{Reference Datasets} To understand the performance of ASR systems on real-world conversational data, we compare them against various standardized datasets. These datasets are commonly used in speech literature to train and benchmark ASR systems. We rely on the following datasets: (i) textbf{LibriSpeech (LS)} is a multilingual dataset derived from audiobooks available on the LibriVox website. Various pre- and post-processing steps have been applied to produce a series of short sequences with accurate transcripts, (ii) \textbf{Fleurs (FL)}: Derived from FLoRes-101, this multilingual dataset consists of Wikipedia sentences narrated by native speakers and then validated by additional workers, (iii) \textbf{CommonVoice (CV)}: Managed by the Mozilla Foundation, this dataset is a large multilingual collection created by volunteers who record their voices by reading a set of sentences. These recordings are then validated by other users to ensure correctness and data quality.

In all these datasets, we rely on the test sets to evaluate the zero-shot performance of ASR models. We report the normalized Word Error Rate for all experiments.


\subsection{Benchmarking ASR over talkbank}

In this experiment, we benchmark various ASR systems on the multilingual TalkBank dataset and compare their performance with other standardized datasets. The results of this experiment are shown in Table~\ref{tab:wer-overall}.

\begin{figure*}[!ht]

    \centering
    \includegraphics[width=1\linewidth]{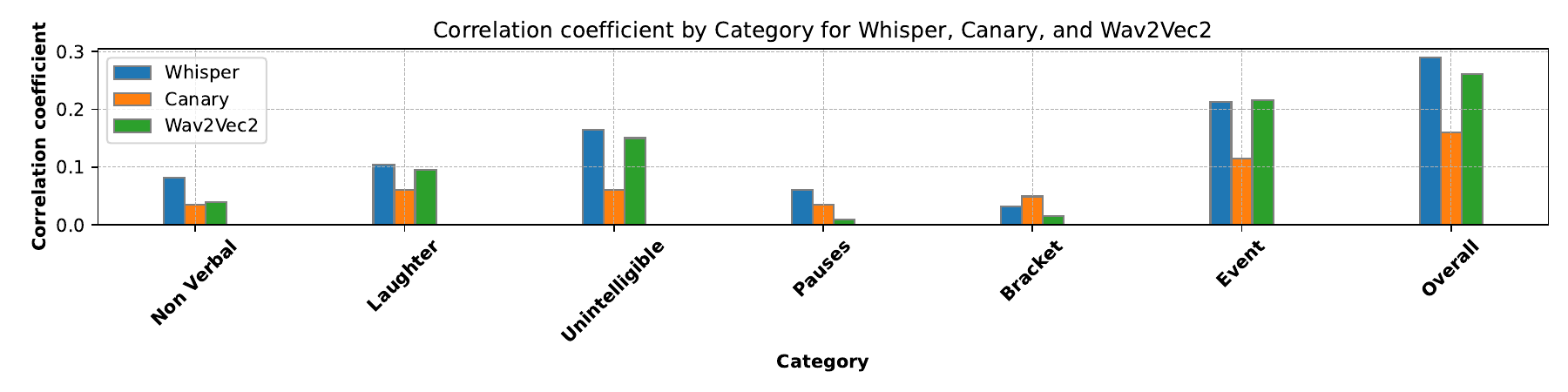}
    \caption{The Pearson correlation coefficient between the number of conversation-specific elements, normalized by the length of the transcript, and WER. Here, 'non-verbal' includes expressions like 'mhh' and 'huh'; 'special characters' refer to speech intonation; 'special utterances' refer to markers like trailing off and interruptions; and 'events' refer to actions such as coughing, groaning, and sneezing. Finally "overall" refers to combined effect.}
    \label{fig:correlation-conv-wer}
\end{figure*}

All ASR models perform significantly worse on the TalkBank dataset compared to LS, FL, and CV dataset. For instance, Whisper exhibits a performance difference that is over five times greater between LS and the TalkBank Segment. Similarly, wav2vec2 shows a twofold performance gap between FL and the TalkBank Segment. These observations suggest that while current ASR model perform well in controlled settings, they struggle in more conversational scenarios.

When comparing the two variants of TalkBank—Segments and Speaker Switch—our results indicate that all approaches perform better on Segments, with the difference being over three times. This can be attributed to the fact that most segments are less than 10 seconds long, with the waveform roughly centered, closely resembling the training files used for ASR systems. In contrast, the Speaker Switch version features audio lengths ranging from 0.1 to 1700 seconds with silences in between, which presents a more challenging setting. Additionally, due to the very long audio file sizes in Speaker Switch, wav2vec2 was unable to process these files, leading us to exclude its results.

In Figure~\ref{fig:wer-language}, we present the results of ASR systems on various languages. Consistent with our earlier findings, all ASR systems exhibit significantly worse performance on the TalkBank dataset compared to others. Unsurprisingly, the lowest WER is observed for English, which we hypothesize is due to the predominance of English data in the training datasets of these ASR models. For other languages, the performance over TalkBank is generally comparable, though slightly worse for French. As before, ASR systems perform better on the TalkBank Segment compared to the Speaker Switch variant.

In Table~\ref{tab:wer-time}, we present the results of ASR systems based on different time durations. Our analysis shows that varying time segments do not significantly affect the Word Error Rate (WER). For example, Whisper has a WER of 0.56 for segments shorter than 5 seconds (T\textless5) and a WER of 0.58 for segments longer than 20 seconds (T\textgreater20). We attribute this robustness to the fact that these ASR models have been primarily trained on short audio segments, which aligns closely with the duration of the TalkBank segments.

\subsection{Effect of conversation specific element and WER}


In this experiment, we explore the impact of conversational-specific markers in audio on the performance of ASR systems. These markers include: (i) non-verbal cues such as utterances like "mhh," "mhm," "umph"; (ii) events like coughing, groaning, and sneezing; (iii) special characters that represent speech intonation changes, such as shifts to high pitch or falling to mid; and (iv) special event terminators that indicate trailing off, interruptions, or questions. A complete list of these markers can be found in the CHAT manual\footnote{\url{https://talkbank.org/manuals/CHAT.html}}. By analyzing these elements, we aim to better understand how ASR systems handle the complexities of natural, conversational speech.

More specifically, in Figure~\ref{fig:correlation-conv-wer}, we plot the correlation score between the number of conversational-specific markers, normalized by the length of the transcripts, and the associated WER. The "overall" category represents the combined effect of all conversational-specific markers. Our findings indicate that all ASR models exhibit a small positive correlation between the presence of these markers and the WER. Notably, events and special characters show a stronger correlation with WER. Interestingly, the models demonstrate robustness to pauses and special utterance terminators, indicating that these factors do not significantly impact ASR performance.

\section{Conclusion}
In this work, we introduced a new conversationally oriented ASR benchmark based on a subset of TalkBank. Our experiments show that modern ASR systems consistently underperform on this dataset compared to more standardized benchmarks, such as LibriSpeech. Additionally, we find a correlation between conversational-specific markers and Word Error Rate (WER). These findings highlight the shortcomings of current ASR models in handling real-world conversational speech and underscore the need for more representative benchmarks.

In the future, we plan to expand the dataset to include gender and other demographic information, release fine-tuned ASR models, and incorporate even more diverse conversational settings to further improve the robustness of ASR systems in practical applications.

\bibliographystyle{IEEEbib}
\bibliography{refs}

\end{document}